\documentclass[runningheads]{llncs}

\usepackage[T1]{fontenc}
\usepackage{graphicx}
\usepackage{hyperref}
\usepackage{subcaption}
\usepackage{subcaption}
\usepackage{multirow} 
\usepackage{graphicx}
\usepackage{booktabs}
\usepackage{cite}
\usepackage{amsfonts} 
\usepackage{amsmath,amssymb}
\usepackage{mathrsfs}
\usepackage{tabularx} 
\usepackage{caption}  
\usepackage{url}
\usepackage{array}
\usepackage{colortbl}
\usepackage[table]{xcolor}
\usepackage{pifont}
\usepackage{float}

\begin{document}

\title{MM-Retinal: Knowledge-Enhanced Foundational Pretraining with Fundus Image-Text Expertise}

\author{Paper ID: $375$}
\author{Ruiqi Wu\inst{1,2} \and
Chenran Zhang\inst{1,2} \and
Jianle Zhang\inst{1,2} \and
Yi Zhou\inst{1,2}\thanks{Corresponding author: Yi Zhou} \and \\
Tao Zhou\inst{3} \and
Huazhu Fu\inst{4}}

\authorrunning{R. Wu and Y. Zhou et al.}

\institute{School of Computer Science and Engineering, Southeast University, China \and Key Laboratory of New Generation Artificial Intelligence Technology and Its Interdisciplinary Applications, Ministry of Education, China \and Nanjing University of Science and Technology, Nanjing, China \and
Agency for Science, Technology and Research (A*STAR), Singapore \\
\email{\{ruiqiwu@seu.edu.cn, yizhou.szcn@gmail.com\}}
}

\maketitle          

\begin{abstract}

Current fundus image analysis models are predominantly built for specific tasks relying on individual datasets. The learning process is usually based on data-driven paradigm without prior knowledge, resulting in poor transferability and generalizability. To address this issue, we propose \textbf{MM-Retinal}, a multi-modal dataset that encompasses high-quality image-text pairs collected from professional fundus diagram books. Moreover, enabled by MM-Retinal, we present a novel \textbf{K}nowledg\textbf{e}-\textbf{e}nhanced foundational \textbf{p}retraining model which incorporates \textbf{F}undus \textbf{I}mage-\textbf{T}ext expertise, called \textbf{KeepFIT}. It is designed with image similarity-guided text revision and mixed training strategy to infuse expert knowledge. Our proposed fundus foundation model achieves state-of-the-art performance across six unseen downstream tasks and holds excellent generalization ability in zero-shot and few-shot scenarios. MM-Retinal and KeepFIT are available at \href{https://github.com/lxirich/MM-Retinal}{https://github.com/lxirich/MM-Retinal}.

\keywords{Fundus image  \and Foundational pretraining \and Knowledge-enhanced}
\end{abstract}

\section{Introduction}
Deep learning has achieved great progress in fundus image analysis. However, most previous works\cite{wu2024medsegdiff,diao2023classification,zhou2023representation,li2021multi} usually utilize individual datasets to train task-specific models. This fashion results in three major model weaknesses: 1) poor generalization ability and robustness across varying scenarios; 2) lack of professional fundus domain-knowledge guidance in learning phase; 3) a huge demand for annotated training data. These challenges underscore the need for developing a general-purpose foundation model which is able to analyze comprehensive ocular diseases in fundus image area. Moreover, learning such a model with less training data but more prior knowledge is preferred. 

In fundus image field, there are many specific image-only public datasets for diverse ocular diseases, such as glaucoma \cite{orlando2020refuge,li2019attention}, diabetic retinopathy\cite{cuadros2009eyepacs,decenciere2014feedback,krause2018grader,zhou2019collaborative}, age-related macular degeneration\cite{ADAM}, and pathological myopia\cite{palm}. Despite remarkable progress of foundation models has been witnessed in many fields, such as natural images\cite{wu2023visual, minigpt4}, medical images like chest X-rays\cite{tiu2022expert, pellegrini2023xplainer} and MRI\cite{lei2023unibrain, liu2023clip}, it lags far behind in fundus image area. Thus, it is of considerable value to explore foundational pretraining in this area. RETFound\cite{zhou2023foundation} and FLAIR\cite{silva2023foundation} made two preliminary attempts but still suffer from certain limitations compared to other foundation models. RETFound only relies on large-scale image data and adopts a masked image modeling manner. FLAIR exploits both vision and language modalities through contrastive pretraining objective, yet it simply maps category label names to fixed texts. Both of them still lacks integrating fundus expertise with rich and profound image-text descriptions.

To accomplish our intention, a high-quality vision-language fundus dataset containing expert knowledge is required. Such a dataset aims to not only promote the development of foundational fundus models with strong generalizability, robustness, and transferability, but also advance research in incorporating knowledge into learning more interpretable models without manual label annotation. Additionally, it should propel research of multimodal fundus image analysis and beyond. Therefore, we built \textbf{MM-Retinal}, a multi-modal dataset which comprises image-text paired data of color fundus photography (CFP), fundus fluorescein angiography (FFA), and optical coherence tomography (OCT) images. All these data are collected from fundus diagram books containing comprehensive ocular knowledge, with accurate image-text descriptions provided by ophthalmologists. Moreover, we also developed \textbf{KeepFIT}, a knowledge-enhanced foundation model, enabled by MM-Retinal. Specifically, image similarity-guided text revision method and mixed training strategy are proposed to inject fundus expert knowledge from our constructed MM-Retinal into model training.

\textbf{We highlight our main contributions:} \textbf{1)} We construct a multi-modal MM-Retinal with over 4.3K high-quality image-text pairs in CFP, FFA and OCT modalities. These pairs are accurately matched with long texts, extensive vocabulary and comprehensive ocular disease and abnormalities. \textbf{2)} A knowledge-enhanced foundation model, KeepFIT, is proposed including a vision-language pre-training framework and knowledge integration methods, termed image similarity-guided text revision and mixed training strategy. \textbf{3)} KeepFIT achieves state-of-the-art performance on six representative downstream tasks, especially in zero-shot and few-shot scenarios, demonstrating consistently strong robustness, generalizability and transferability.

\section{The MM-Retinal Dataset}
\subsection{Dataset Construction}
\label{sec:data_collection}

\begin{figure*}[t]
    \centering
    \includegraphics[width=\linewidth]{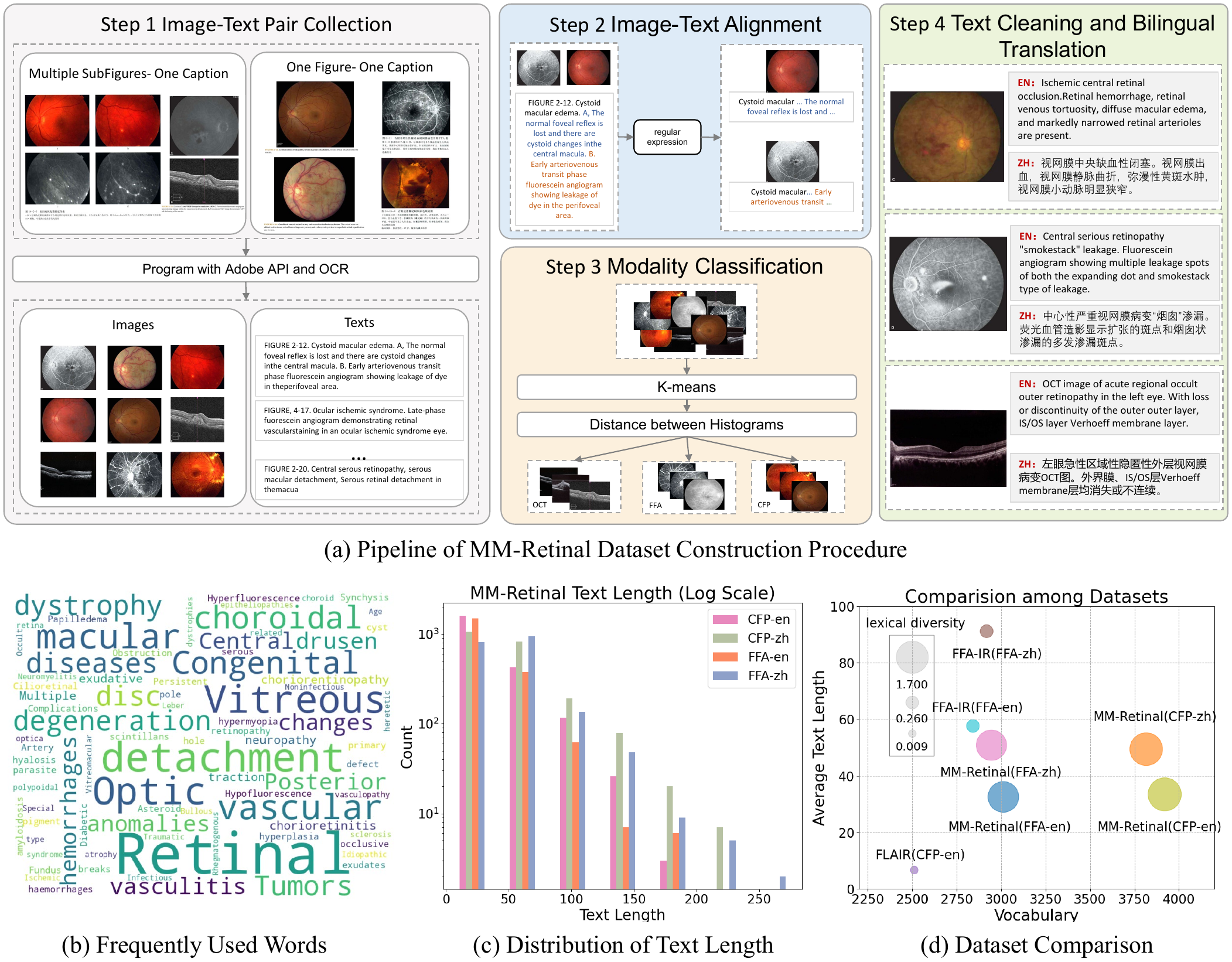}
    \caption{Construction workflow and statistical overview of MM-Retinal.}
    \vspace{-1ex}
    \label{fig:collection}
\end{figure*}

To construct MM-retinal with high-quality image-text pairs covering CFP, FFA, and OCT modalities, as shown in Fig.~\ref{fig:collection}(a), our designed semi-automatic pipeline of dataset construction contains four steps. For further details, please refer to the supplementary material. A six-person team took four weeks to get MM-Retinal completed.

\textbf{Step 1: Image-Text Pair Collection}: First, image-text pairs are captured from the books. Note that we keep the resolution of each image in the pair no less than 800×800. Afterwards, Adobe API and OCR techniques are used for raw image and text extraction. \textbf{Step 2: Image-Text Alignment}: As mentioned earlier, not all the initially extracted images and texts by the program are very well aligned, as there are some sub-figures correspond to one caption within a screenshot. Thus, we use regular expression matching for sub-figure caption separation to align image-text pairs. \textbf{Step 3: Modality Classification}: Since the fundus images in the books include multiple modalities, we categorize images into CFP, FFA, and OCT using K-means and color histogram analysis, and excluding unusual modalities with scarce samples. \textbf{Step 4: Text Cleaning and Bilingual Translation}: After manually correcting OCR errors and irrelevant texts, we translate the texts to provide bilingual (English and Chinese) versions, aiming to standardize the language and enhance the influence of MM-Retinal since the diagram books we used include both Chinese and English. 
\subsection{Dataset Statistics}

\label{sec:data_statistics}
Current version of MM-Retinal dataset includes 2,169 CFP cases, 1,947 FFA cases and 233 OCT cases. Each case is provided with an image and texts in both English and Chinese. Due to the small scale of OCT modality, we do not explore it for now and will extend this part in future. Detailed statistical analysis of the CFP and FFA data is provided in the Fig. ~\ref{fig:collection}, focusing on aspects of frequently used words, text length, vocabulary size, and comparison with other datasets.

\textbf{Frequently Used Words:} Since MM-Retinal dataset is built based on comprehensive ocular fundus diagram books, it covers a wide range of disease categories, such as macular, retinal vascular, and optic nerve diseases. In Fig. ~\ref{fig:collection}(b), we only show a small part of the words frequently appeared in CFP and FFA modalities. More retinal abnormal fundus changes and disease categories can be found in the supplementary material.

\textbf{Text Length:} Fig.~\ref{fig:collection}(c) illustrates the text length distribution of our dataset. About 75\% of English texts and 45\% of Chinese texts range from 1 to 40 words, while 19\% of English texts and 43\% of Chinese texts contain between 41 and 80 words. Since our dataset is sourced from ophthalmologists’ diagram books, it features much longer texts compared to FLAIR which simply maps class names of public datasets into fixed texts.

\textbf{Vocabulary Size:} MM-Retinal dataset contains diverse textual descriptions, such as disease diagnosis, lesion characteristics (e.g. color, shape, appearance), clinical manifestations, and post-treatment efficacy information. Fig.~\ref{fig:collection}(d) showcases the average vocabulary size and lexical diversity across different modalities and languages, where average vocabulary size represents the total number of unique words contained in the whole texts within the dataset and lexical diversity refers to the vocabulary size averaged over each text.

\textbf{Discussion:} Compared to public fundus image datasets, our MM-Retinal stands out from four aspects: \textbf{1) Multi-modality:} MM-Retinal is the pioneering fundus dataset that includes high-quality image-text expertise data for CFP, FFA, and OCT. \textbf{2) Data Quality.} In contrast to other medical image datasets that often contain low-quality images and captions sourced from websites or academic papers, MM-Retinal provides high-quality images with a resolution over $800\times 800$, accompanied by accurate and pertinent text descriptions. The images closely match clinical data with minimal domain shift. \textbf{3) Category Variety.} According to the contents of four diagram books, MM-Retinal covers a broad range of categories with over 96 abnormalities and diseases. \textbf{4) Text Diversity.} MM-Retinal encompasses diverse vocabulary and long texts, containing extensive expert knowledge, and can be explored to enhance data-driven fundus image analysis models.

\section{Knowledge-Enhanced Foundational Pretraining}
\subsection{Vision-Language Pre-training Framework}
\label{sec:architecture}
\begin{figure*}[t]
    \centering
    \includegraphics[scale=0.4]{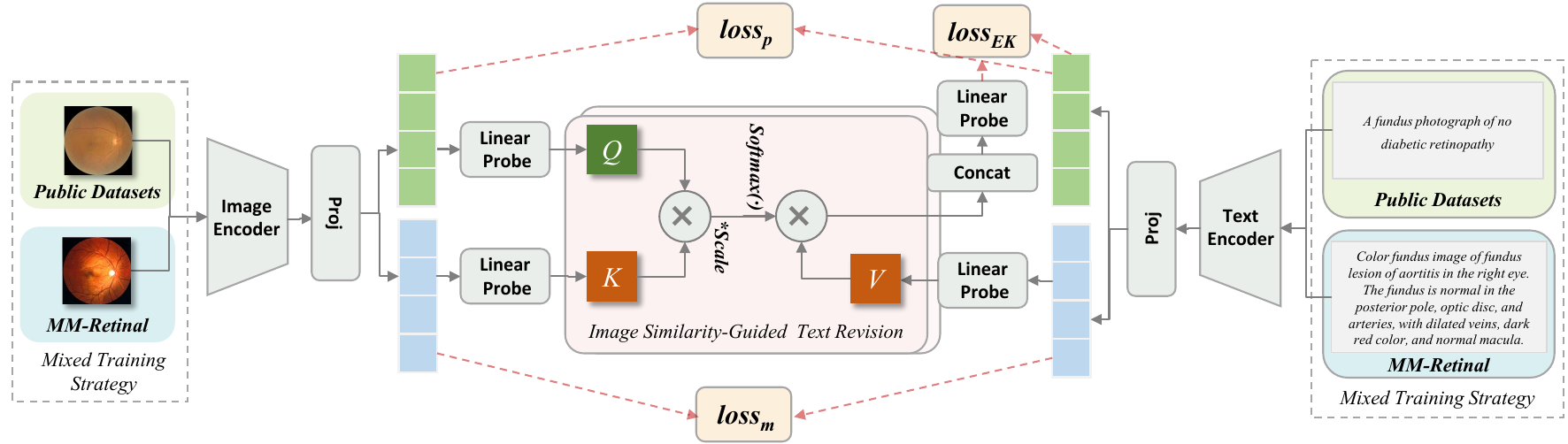}
    \caption{\textbf{KeepFIT}: A vision-language pretraining framework using image-guided text revision and mixed training methods to infuse expert knowledge.}
    \label{fig:keepfit}
\vspace{-1ex}
\end{figure*}

As shown in Fig. ~\ref{fig:keepfit}, we propose \textbf{KeepFIT}, a knowledge-enhanced foundational model, pretrained on public and MM-Retinal datasets. In this paper, we define public datasets with only  category-level labels as unimodal dataset and we follow FLAIR\cite{silva2023foundation} to fill category-level labels into a prompt template to create texts. We utilize ResNet50\cite{he2016deep} as image encoder $E_v$ and BioClinicalBert\cite{alsentzer2019publicly} as text encoder $E_t$, followed by projection heads $P_v$ and $P_t$ to match image and text feature dimensions $d$. Thus, given a image $x_i$ and a text $y_j$, the extracted image features $v_i$ and text features $t_j$ are:
\begin{equation}
    v_i = P_{v} \circ E_{v}(x_i) \in \mathbb{R^{d}}^{d},~~ t_j = P_{t} \circ E_{t}(y_j) \in \mathbb{R^{d}}^{d}.
\end{equation}

Given that texts detail the associated fundus images, our goal is to maximize similarity between paired image-text and minimize similarity for unpaired ones in a multimodal space. For MM-retinal, which are genuine texts rather than textual prompts, we follow CLIP\cite{radford2021learning} to implement the contrastive loss as shown in Eq.~\ref{mloss}, and the matching labels $G_{v2t}^{m}$ and $G_{t2v}^{m}$ are two $\mid \mathcal{B} \mid \times \mid \mathcal{B} \mid$ identity matrices:

\begin{equation}
\label{mloss}
\mathcal{L}_{m} =\mathbb{E}_{(x,y)\sim\mathcal{B}}[CE(G_{v2t}^{m},S_{v2t}) + CE(G_{t2v}^{m},S_{t2v})],   G_{v2t}^{m}, G_{t2v}^{m} \in I_{\mid \mathcal{B} \mid},
\end{equation}

\noindent{where $m$ represents our proposed MM-Retinal, $\mathcal{B}$ is the batchsize, $CE$ denotes InfoNCE loss, $S_{v2t} = \lambda(v_i {t_j}^\top)$ and $S_{t2v} = \lambda(v_i^\top {t_j})$ are the cross-modality similarity, $\lambda$ is a learnable scaling factor, $v2t$ and $t2v$ denote image-to-text and text-to-image, respectively.}

For public datasets that only have prompt texts mapped from category-level labels, we follow FLAIR to calculate the category co-occurrence relationships among samples within a batch and construct a target matrix to encourage the pairs belonging to the same category closer. Thus, the objective function Eq.~\ref{mloss} is converted into:
\begin{equation}
\label{uloss}
\mathcal{L}_{p} =\mathbb{E}_{(x,y)\sim\mathcal{B}}[CE(G_{v2t}^{p},S_{v2t}) + CE(G_{t2v}^{p},S_{t2v})],
\end{equation}
\begin{equation}
\label{matching}
G^p_{{v}2{t}} = G^p_{{t}2{v}} = \begin{cases} 
1, & \text{if } \text{category}_{v} = \text{category}_{t} \\
0, & \text{otherwise}
\end{cases}
\end{equation}

\noindent{where $p$ represents public datasets that only have category-level labels, the matching labels $G_{v2t}^{p}, G_{t2v}^{p}$ are $\mid \mathcal{B} \mid \times \mid \mathcal{B} \mid$ symmetric matrices}.

\subsection{Expert Knowledge Integration Methods}
\label{sec:knowledge}

As our proposed MM-Retinal features high-quality image-text pairs, long text description, rich vocabulary, and comprehensive ocular diseases categories, it encapsulates extensive fundus expert knowledge. Inspired by TipAdapter\cite{zhang2022tip}, we propose a lightweight expert knowledge integration method, called \textbf{Image Similarity-Guided Text Revision} along with \textbf{Mixed Training Strategy}, to boost the expert knowledge from our MM-Retinal dataset.

{\textbf{Image Similarity-Guided Text Revision:}} The descriptions within MM-Retinal are more comprehensive and professional than the simple texts mapped from class names in public datasets. Despite this, the images from both sources share a notable similarity. Hence, we start from identifying visual features in the public datasets that resemble those in MM-Retinal. Visual similarities are used as guidance to extract relevant prior knowledge from MM-Retinal's text features to refine and enhance the textual prompts of the public datasets. 

Specifically, given an input image-text pair $[x_p, y_p]$ from public datasets and $[x_m, y_m]$ from MM-Retinal, the extracted features are $(v_p, t_p)$ and $(v_m, t_m)$. A multi-head cross-attention\cite{vaswani2017attention} is applied to exploit prior expert knowledge as:
\begin{equation}
\label{MHA}
EK = \text{MultiHead}(Q, K, V) = \text{Concat}(\text{head}_1, \text{head}_2,\ldots,\text{head}_h)W^O,
\end{equation}
\begin{equation}
\label{ATTN}
\text{head}_i = \text{ATTN}(v_pW^Q_i, v_mW^K_i, t_mW^V_i),
\end{equation}

\noindent{where $W^Q_i, W^K_i, W^V_i$ and $W^O$ are parameter matrics for projection, $v_p, v_m, t_m$ refer to image features of public datasets, image features of MM-Retinal and text features of MM-Retinal, respectively.} Then, we establish an expert knowledge revision loss based on Mean Squared Error (MSE) to refine the text features of public datasets by incorporating expert knowledge, as formulated in Eq.~\ref{MESloss}:
\begin{equation}
\label{MESloss}
\mathcal{L}_{EK} = \text{MSELoss}(EK, t_p).
\end{equation}

{\textbf{Mixed Training Strategy:}} Since the public datasets has relatively homogeneous text prompts and almost no expert knowledge, which is vastly different from the texts of our dataset, we propose a mixed training strategy to avoid model optimization bias during the training process. To detail, the samples from public datasets and our dataset are in a 1:1 ratio in each batch.

{\textbf{Overall Training Objective:}} The overall training objective comprises three parts, which are public datasets contrastive loss $\mathcal{L}_{p}$, MM-Retinal contrastive loss $\mathcal{L}_{m}$ and expert knowledge revision loss $\mathcal{L}_{EK}$:
\begin{equation}
\label{loss}
\mathcal{L} = \mathcal{L}_{p} + \mathcal{L}_{m} + \alpha \mathcal{L}_{EK},
\end{equation}

\noindent{where $\alpha$ is to weight for $\mathcal{L}_{EK}$, empirically set as 100 showing best performance.}

\section{Experiment and Results}
\subsection{Datasets and Baselines}
\label{sec:baselines}

\textbf{Pre-training Data}: \textbf{1) flair\cite{silva2023foundation} (CFP)} compiles 37 open-access fundus image datasets covering 96 categories with up to 284,660 images. These datasets provide category-level labels for classification.
\textbf{2) SYNFUNDUS-1M\cite{shang2023synfundus} (CFP)} is a synthetic dataset with 1 million images for 14 diseases, created by a diffusion model\cite{ho2020denoising} trained on 1.3 million private fundus images.
\textbf{3) FFA-IR\cite{li2021ffa} (FFA)} provides 10,790 reports along with 1,048,584 images from clinical practice. It includes a schema of 46 categories of lesion and bilingual reports.
\textbf{4) MM-Retinal (CFP+FFA+OCT)} contains over 4.3K high-quality image-text pairs from professional fundus diagram books, covering 96 categories of abnormalities and diseases.

\begin{table}[t]
\centering
\caption{Comparison of generalization ability on Few-Shot and Zero-Shot tasks. FLAIR and flair denote the model and datasets in \cite{silva2023foundation}, respectively. SynFuduns-1M abbreviated as syn. MM represents MM-Retinal.}
\fontsize{5}{10.8}\selectfont
\renewcommand\arraystretch{0.8}
\begin{tabular}{c|c|ccccccccc|ccc|c}
\toprule
\multirow{5}{*}{Method}  & \multirow{5}{*}{Data} & \multicolumn{9}{c|}{Few-Shot}                                                                        & \multicolumn{3}{c|}{Zero-Shot}                                     &                      \\ \cline{3-14} 
                         &                       & \multicolumn{9}{c|}{ODIR200×3}                                                                       & REFUGE               & FIVES                & ODIR200×3            & \multirow{4}{*}{Avg} \\ \cline{3-14}
                         &                       & \multicolumn{3}{c}{ClipAdapter} & \multicolumn{3}{c}{TipAdapter} & \multicolumn{3}{c|}{TipAdapter-f} & \multirow{3}{*}{AUC} & \multirow{3}{*}{ACA} & \multirow{3}{*}{ACA} &                      \\
                         &                       & 1        & 5        & 10        & 1        & 5        & 10       & 1         & 5         & 10        &                      &                      &                      &                      \\
                         &                       & \multicolumn{3}{c}{ACA}         & \multicolumn{3}{c}{ACA}        & \multicolumn{3}{c|}{ACA}          &                      &                      &                      &                      \\ \midrule
\multirow{2}{*}{FLAIR}   & flair                 & 0.720         & 0.823         & 0.863          & 0.403         & 0.413         & 0.422         & 0.417          & 0.462          & 0.535          & 0.926                & 0.670                & 0.403    & 0.588                     \\
                         & flair+syn             & 0.735         & 0.827         & 0.852          & 0.603         & 0.622         & 0.632         & 0.580          & 0.647          & 0.672          & 0.880                & 0.617                & 0.520    & 0.682                     \\ \midrule
\multirow{4}{*}{KeepFIT} & flair                 & 0.763        & 0.848         & 0.847          & 0.780         & 0.782         & 0.795         & 0.775          & 0.785          & 0.803          & 0.931                     & 0.666                     & 0.768                     & 0.795                     \\
                         & flair+syn             & 0.795         & 0.858         & 0.862          & 0.751         & 0.777         & 0.783         & 0.760          & 0.795          & 0.807          & 0.856                & 0.696                & 0.777     & 0.793                     \\
                         & 50\%flair+MM          & 0.832     & 0.873    & 0.887    & 0.862    & 0.870    & 0.872    & 0.870     & 0.883     & 0.873     & 0.934                & 0.654                & 0.862                & \cellcolor{gray!60}0.856                      \\
                         & flair+MM              & 0.848         & 0.878         & 0.893          & 0.823         & 0.843         & 0.842         & 0.820          & 0.847          & 0.853          & 0.941                & 0.731                & 0.812     & \cellcolor{gray!30}0.844                     \\
\bottomrule
\end{tabular}
\label{table:zeroshot}
\end{table}

\textbf{Foundation Model Baselines}: \textbf{1) MoE\cite{wang2019retinal} (MTL)} uses a multi-task approach with mixture-of-experts for fundus, macula, and optic disc images.
\textbf{2) RETFound\cite{zhou2023foundation} (MIM)} is a masked autoencoder with Transformers\cite{dosovitskiy2020vit}, trained for retinal image reconstruction.
\textbf{3) FLAIR\cite{silva2023foundation} (CLIP)} utilizes a CLIP model and brief textual prompts that mapped from category labels for pre-training, but the texts are so limited and brief that fail to fully describe the images.

\textbf{Evaluation Data and Metrics}: For CFP modality, we evaluate on five datasets across finetuning, few-shot, and zero-shot settings, using classification accuracy (ACA)\cite{zhao2019bira} for REFUGE\cite{orlando2020refuge}, FIVES\cite{jin2022fives}, ODIR200$\times$3\cite{odir2019} and TAOP\cite{taop2021}, receiving-operative-curve(AUC) for REFUGE and AMD\cite{ADAM}, F1-score for AMD. For FFA, we evaluate on image captioning task using the FFA-IR test split, applying BLUE 1-4, Meter, Rouge, and Cider metrics for consistency with FFA-IR.

\textbf{Implementation Summary}: In CFP and FFA, image and text encoders are initialized from ImageNet-1K\cite{russakovsky2015imagenet} and BioClinicalBERT, respectively. Attention mechanisms are trained from scratch, with 512 feature dimensions. Images are adjusted to $512\times512$ size. All the texts we used are in English version. Text tokens length is set at 256. Evaluation uses five-fold cross-validation averaging. We employ AdamW (lr=1e-4, decay=1e-5) optimizer and cosine scheduler with initial warm-up for the first epoch. Training is conducted on 4 RTX 3090 GPUs with batches of 24.

\subsection{Comparison of Generalization Ability on Zero-Shot and Few-Shot Tasks}
\label{sec:zero_shot}

\begin{table}[t]
\caption{Comparison of transferability on unseen downstream datasets and ablation study. FT refers to finetune.}
\fontsize{5}{12.8}\selectfont
\renewcommand\arraystretch{0.8}
\begin{tabular}{ccccccccccccc}
\toprule
\multicolumn{13}{c}{
  \begin{minipage}[c][6.5ex][c]{\linewidth}
    \scriptsize \textbf{(a) Comparison of transferability on unseen downstream datasets(CFP)}
  \end{minipage}
} \\ \toprule                                                                                                                                                                                        
\multicolumn{1}{c|}{\multirow{3}{*}{Type}} & \multicolumn{1}{c|}{\multirow{3}{*}{Method}}  & \multicolumn{1}{c|}{\multirow{3}{*}{Data}} & \multicolumn{1}{c|}{\multirow{3}{*}{Data Size}} & \multicolumn{3}{c|}{FT(20\%-20\%)}         & \multicolumn{1}{c|}{FT(train-val)} & \multicolumn{2}{c|}{FT(80\%-20\%)} & \multirow{3}{*}{Avg} \\ \cline{5-10}
\multicolumn{1}{c|}{}                      & \multicolumn{1}{c|}{}                         & \multicolumn{1}{c|}{}                      & \multicolumn{1}{c|}{}                           & REFUGE & FIVES  & \multicolumn{1}{c|}{ODIR200×3} & \multicolumn{1}{c|}{TAOP}                  & \multicolumn{2}{c|}{AMD}                 &                      \\
\multicolumn{1}{c|}{}                      & \multicolumn{1}{c|}{}                         & \multicolumn{1}{c|}{}                      & \multicolumn{1}{c|}{}                           & ACA    & ACA    & \multicolumn{1}{c|}{ACA}       & \multicolumn{1}{c|}{ACA}                   & AUC      & \multicolumn{1}{c|}{F1}       &                      \\ \midrule
\multicolumn{1}{c|}{MTL}                   & \multicolumn{1}{c|}{MoE}                      & \multicolumn{1}{c|}{flair}                 & \multicolumn{1}{c|}{278,348}                    & 0.543  & 0.364  & \multicolumn{1}{c|}{0.609}     & \multicolumn{1}{c|}{0.239}                 & 0.539    & \multicolumn{1}{c|}{0.405}    & 0.450                \\ \midrule
\multicolumn{1}{c|}{MIM}                   & \multicolumn{1}{c|}{RETFound}                 & \multicolumn{1}{c|}{RETFound}              & \multicolumn{1}{c|}{904,170}                    & 0.809  & 0.765  & \multicolumn{1}{c|}{0.907}     & \multicolumn{1}{c|}{0.697}                 & 0.945    & \multicolumn{1}{c|}{0.805}    & 0.821                \\ \midrule
\multicolumn{1}{c|}{\multirow{6}{*}{CLIP}} & \multicolumn{1}{c|}{\multirow{2}{*}{FLAIR}}   & \multicolumn{1}{c|}{flair}                 & \multicolumn{1}{c|}{278,348}                    & 0.831  & 0.835  & \multicolumn{1}{c|}{0.875}     & \multicolumn{1}{c|}{0.468}                 & 0.963    & \multicolumn{1}{c|}{0.960}    & 0.822                \\
\multicolumn{1}{c|}{}                      & \multicolumn{1}{c|}{}                         & \multicolumn{1}{c|}{flair+syn}             & \multicolumn{1}{c|}{1,278,366}                  & 0.847  & 0.842  & \multicolumn{1}{c|}{0.890}     & \multicolumn{1}{c|}{0.549}                 & 0.952    & \multicolumn{1}{c|}{0.953}    & 0.839                \\ \cline{2-11} 
\multicolumn{1}{c|}{}                      & \multicolumn{1}{c|}{\multirow{4}{*}{KeepFIT}} & \multicolumn{1}{c|}{flair}                 & \multicolumn{1}{c|}{278,348}                    & 0.837  & 0.838  & \multicolumn{1}{c|}{0.903}     & \multicolumn{1}{c|}{0.556}                 & 0.951    & \multicolumn{1}{c|}{0.957}    & 0.840                \\
\multicolumn{1}{c|}{}                      & \multicolumn{1}{c|}{}                         & \multicolumn{1}{c|}{flair+syn}             & \multicolumn{1}{c|}{1,278,366}                  & 0.832  & 0.842  & \multicolumn{1}{c|}{0.890}     & \multicolumn{1}{c|}{0.579}                 & 0.976    & \multicolumn{1}{c|}{0.969}    & 0.848                \\
\multicolumn{1}{c|}{}                      & \multicolumn{1}{c|}{}                         & \multicolumn{1}{c|}{50\%flair+MM}          & \multicolumn{1}{c|}{141,343}                    & 0.856  & 0.834  & \multicolumn{1}{c|}{0.913}     & \multicolumn{1}{c|}{0.700}                 & 0.962    & \multicolumn{1}{c|}{0.962}    & \cellcolor{gray!30}0.871                \\
\multicolumn{1}{c|}{}                      & \multicolumn{1}{c|}{}                         & \multicolumn{1}{c|}{flair+MM}              & \multicolumn{1}{c|}{280,517}                    & 0.861  & 0.851  & \multicolumn{1}{c|}{0.915}     & \multicolumn{1}{c|}{0.684}                 & 0.971    & \multicolumn{1}{c|}{0.966}    & \cellcolor{gray!60}0.875                \\ \bottomrule
\multicolumn{13}{c}{
  \begin{minipage}[c][6.5ex][c]{\linewidth}
    \scriptsize \textbf{(b) Comparison of transferability on unseen downstream datasets(FFA)}
  \end{minipage}
} \\ \toprule                                                                                                                                                                                        
\multicolumn{1}{c|}{Model}                 & \multicolumn{1}{c|}{Data}                     & \multicolumn{1}{c|}{Data Size}             & B1                                              & B2     & B3     & B4                             & Meter                                      & Rouge    & \multicolumn{1}{c|}{Cider}    & Avg                  \\ \midrule
\multicolumn{1}{c|}{CNN + T}               & \multicolumn{1}{c|}{FFA-IR}                   & \multicolumn{1}{c|}{1,048,584}             & 0.321                                           & 0.211  & 0.154  & 0.122                          & 0.198                                      & 0.268    & \multicolumn{1}{c|}{0.283}    & 0.222                \\ \midrule
\multicolumn{1}{c|}{CNN + T}               & \multicolumn{1}{c|}{FFA-IR+MM}                & \multicolumn{1}{c|}{1,050,531}             & 0.363                                           & 0.244 & 0.171 & 0.127                         & 0.149                                     & 0.302   & \multicolumn{1}{c|}{0.314}   & \cellcolor{gray!60}0.239                \\ \bottomrule
\multicolumn{13}{c}{
  \begin{minipage}[c][6.5ex][c]{\linewidth}
    \scriptsize \textbf{(c) Ablation Study(MHCA denotes multi-head cross attention)}
  \end{minipage}
} \\ \toprule                                                                                                                                                                                        
\multicolumn{1}{c|}{\multirow{3}{*}{Model}}   & \multicolumn{1}{c|}{Revision}                   & \multicolumn{1}{c|}{Mixed}                 & \multicolumn{1}{c|}{Fusion}                     & \multicolumn{3}{c|}{FT(20\%-20\%)}         & \multicolumn{1}{c|}{FT(train-val)} & \multicolumn{2}{c|}{FT(80\%-20\%)} & \multirow{3}{*}{Avg} \\ \cline{2-10}
\multicolumn{1}{c|}{}                         & \multicolumn{1}{c|}{\multirow{2}{*}{MHCA}}    & \multicolumn{1}{c|}{\multirow{2}{*}{/}}    & \multicolumn{1}{c|}{\multirow{2}{*}{MHCA}}      & REFUGE & FIVES  & \multicolumn{1}{c|}{ODIR200×3} & \multicolumn{1}{c|}{TAOP}                  & \multicolumn{2}{c|}{AMD}                 &                      \\
\multicolumn{1}{c|}{}                         & \multicolumn{1}{c|}{}                         & \multicolumn{1}{c|}{}                      & \multicolumn{1}{c|}{}                           & ACA    & ACA    & \multicolumn{1}{c|}{ACA}       & \multicolumn{1}{c|}{ACA}                   & AUC      & \multicolumn{1}{c|}{F1}       &                      \\ \hline
\multicolumn{1}{c|}{\multirow{5}{*}{KeepFIT}} & \multicolumn{1}{c|}{{\scalebox{2}{\checkmark}}}                        & \multicolumn{1}{c|}{}                      & \multicolumn{1}{c|}{}                           & 0.803  & 0.841  & \multicolumn{1}{c|}{0.903}     & \multicolumn{1}{c|}{0.640}                 & 0.964    & \multicolumn{1}{c|}{0.884}    & 0.839                \\
\multicolumn{1}{c|}{}                         & \multicolumn{1}{c|}{}                         & \multicolumn{1}{c|}{{\scalebox{2}{\checkmark}}}                     & \multicolumn{1}{c|}{}                           & 0.832  & 0.851  & \multicolumn{1}{c|}{0.907}     & \multicolumn{1}{c|}{0.675}                 & 0.965    & \multicolumn{1}{c|}{0.905}    & 0.856                \\
\multicolumn{1}{c|}{}                         & \multicolumn{1}{c|}{}                         & \multicolumn{1}{c|}{{\scalebox{2}{\checkmark}}}                     & \multicolumn{1}{c|}{{\scalebox{2}{\checkmark}}}                          & 0.832  & 0.852  & \multicolumn{1}{c|}{0.900}     & \multicolumn{1}{c|}{0.673}                 & 0.963    & \multicolumn{1}{c|}{0.863}    & 0.847                \\
\multicolumn{1}{c|}{}                         & \multicolumn{1}{c|}{{\scalebox{2}{\checkmark}}}                        & \multicolumn{1}{c|}{{\scalebox{2}{\checkmark}}}                     & \multicolumn{1}{c|}{{\scalebox{2}{\checkmark}}}                          & 0.822  & 0.851  & \multicolumn{1}{c|}{0.905}     & \multicolumn{1}{c|}{0.694}                 & 0.964    & \multicolumn{1}{c|}{0.966}    & \cellcolor{gray!30}0.867                \\
\multicolumn{1}{c|}{}                         & \multicolumn{1}{c|}{(Ours){\scalebox{2}{\checkmark}}}                  & \multicolumn{1}{c|}{{\scalebox{2}{\checkmark}}}                     & \multicolumn{1}{c|}{}                           & 0.861  & 0.851  & \multicolumn{1}{c|}{0.915}     & \multicolumn{1}{c|}{0.684}                 & 0.971    & \multicolumn{1}{c|}{0.966}    & \cellcolor{gray!60}0.875                \\ \hline

\end{tabular}
\label{table:foundation}
\end{table}

\label{table:zero_shot}
To compare the foundational performance of different models, we conducted experiments in zero-shot and few-shot scenarios with unseen categories. We adopted 1, 5, 10 shots and adapter tuning scheme in few-shot, including ClipAdapter\cite{gao2024clip}, TipAdapter\cite{zhang2022tip} that adds a few task-specific parameters. 

In Tab~\ref{table:zeroshot}, KeepFIT trained by MM-Retinal and 50\% flair achieves competitive performance across the board. 50\% flair is obtained by randomly sampling 50\% of each dataset in the flair. For example, it outperforms FLAIR trained by flair by 0.268 and KeepFIT trained by flair by 0.061 in average, and even shows superior performance to KeepFIT trained by MM-Retinal and flair with an improvement of 0.012. The results demonstrate several key insights: 1) The best performance comes from using 50\% flair with MM-Retinal, which is even higher than using 100\% flair with MM-Retinal. This indicates that large datasets may introduce noise and diminish transfer effectiveness since they lack expert knowledge. 2) KeepFIT performs better when trained by MM-Retinal and flair than by synfundus-1M and fliar, underscoring MM-Retinal's superior expert knowledge over large datasets like synfundus-1M for model generalization ability and transferability.

\subsection{Comparison of Transferability on Unseen Downstream Datasets}
\label{sec:foundation}

To assess KeepFIT's transferability, we fine-tuned it on six unseen datasets. Specifically for CFP, we added and fine-tuned a fully connected layer to the image encoder and keep the other parts frozen. We conducted five fine-tuning settings, including 20\%, 40\%, 60\%, and 80\% of the data for training, the rest 20\% for testing or following the official dataset partition.  For FFA, due to dataset scarcity, we utilized image captioning for assessment following FFA-IR\cite{li2021ffa}, using ResNet to extract images features and Transformer-based decoder for caption generation. 

As shown in Tab~\ref{table:foundation}(a), KeepFIT trained on MM-Retinal and flair achieves SOTA in almost all the unseen downstream datasets. Especially, it outperforms KeepFIT trained on flair by 0.035 and FLAIR trained on flair by 0.053. Similarly in Tab~\ref{table:foundation}(b), the performance of FFA modality is improved when trained on both MM-Retinal and FFA-IR by 0.017. More results are provided in the supplementary material.

\subsection{Ablation Study}
\label{sec:ablation}

We ablated KeepFIT on all flair and MM-Retinal from three aspects. 1) Image similarity-guided text revision module: assess the necessity of revising the text features of public datasets using the knowledge extracted from MM-Retinal by the guidance of image similarity. 2) Mixed training strategy: test the necessity of including data from two sources in one batch. 3) Text fusion module: we substituted text revision with a text fusion module. It integrates knowledge extracted from MM-Retinal by multi-head cross attention into public dataset text features via residual connections to test the necessity of augmenting text features. 

Table~\ref{table:foundation}(c) shows that image similarity-guided text revision module and mixed training strategy are essential for the performance improvement. This is because revision module injects the appropriate expert knowledge into the texts and mixed training strategy significantly boosts the performance by reducing conflicts. However, the text fusion module makes minimal contribution, suggesting that text revision is more effective at injecting knowledge than text fusion.

\section{Conclusion}
In this work, we built a multi-modal MM-Retinal dataset, with high-quality fundus image-text expertise. We also proposed KeepFIT, a vision-language pretraining framework enhancing expert knowledge infusion. Experimental results highlight its transferability to unseen datasets and generalization ability on few-shot and zero-shot scenarios. We expect this work will open up unexplored topics in fundus research, such as building multimodal knowledge graphs of fundus images, and high-quality text-to-image generation.

\section{Acknowledgement}

This work was partially supported by the National Natural Science Foundation of China (Grants No 62106043, 62172228), and the Natural Science Foundation of Jiangsu Province (Grants No BK20210225). We're also deeply grateful to Yu-Ang Yao, Minqi Gao, Junkai Chen, Jiaqi Li, Zimeng Zhu, and Jiaqi Xu, who have been instrumental in the construction of the MM-Retinal dataset.

\bibliographystyle{splncs04}
\bibliography{ref.bib}

\title{Supp: MM-Retinal: Knowledge-Enhanced Foundational Pretraining with Fundus Image-Text Expertise}
\author{}
\author{}
\institute{}

\maketitle            

\section{Details of MM-Retinal Dataset Construction}

To advance multi-modal fundus foundation model research and foster expert knowledge integration in learning fundus image analysis models, we build MM-Retinal, a high-quality image-text paired fundus dataset that comprises CFP, FFA, and OCT modalities. We design a semi-automatic collection procedure to improve construction efficiency, which consists of four steps: 1) image-text pair collection; 2) image-text alignment; 3) modality classification; 4) text cleaning and bilingual translation.

\textbf{Step 1: Image-Text Pair Collection}. Textual reports for fundus diagnosis are typically not accompanied with images in clinical process. Therefore, unlike X-ray and CT images, it is challenging to directly obtain image-text pairs from fundus clinical reports. To address this, we collect image-text pairs from four diagram books illustrating ocular fundus diseases with high-quality expert captions. First, image-text pairs are captured from the books. In cases that one figure corresponds to one caption, we simply capture them in one screenshot. For the cases that multiple sub-figures correspond to a piece of caption, both the text and its corresponding sub-figures are captured together, as shown in Fig. ~\ref{fig:collection}(a). Note that we keep the resolution of each image in the pair no less than 800×800. Afterwards, these screenshots are filled into the program we designed to parse images by Adobe API and extract texts by OCR technique. Additionally, the color of a certain book suffer a dark tone. We implemented a dehazing operation based on Gamma transformation to correct the color of this book to match the color distribution of other books.

\textbf{Step 2: Image-Text Alignment}. As mentioned earlier, not all the initially extracted images and texts by the program are very well aligned, as there are some sub-figures correspond to one caption within a screenshot. Thus, for those failure cases, we use regular expression matching to split them. Specifically, if the text matches "Figure No.", it indicates the beginning of the text of a new image-text pair. If it matches "Letter.", it indicates the beginning of the text of a new subfigure-text pair. As for the separation of sub-figures, we apply Adobe API to automatically implement subgraph segmentation.

\textbf{Step 3: Modality Classification}. Since the fundus images in the books include multiple modalities, we classified them into CFP, FFA, and OCT, separately, which are the mainstream modalities in fundus domain, and exclude others due to their limited samples. Specifically, we first employ K-means to categorize images into three categories based on their color histograms. It effectively separates the CFP modality which characterized by the distinct color pattern. The other two categories are the mixture of FFA, OCT and images from other modalities. Subsequently, we select a reference image from both OCT and FFA respectively. The classification of remaining images is based on the distance between their color histograms and the references' color histograms. This process results in a precise separation of FFA and OCT modality images.

\textbf{Step 4: Text Cleaning and Bilingual Translation}. The text extraction may have some OCR recognition errors or incomplete issues, so we correct the texts manually. Moreover, to enhance the relevance within an image-text pair, we remove irrelevant information from the text, such as the index of the corresponding image. We also make modifications to sentence inaccuracies. For example, the original text describes multi subgraphs (e.g. `both eyes show'), but after separating them, the text description should be appropriately adjusted to reflect details about each individual subgraph (e.g. `left eye shows' and `right eye shows')". In addition, since the
diagram books we selected include both Chinese and English, we provide bilingual reports in English and Chinese version to standardize the language and make MM-Retinal more influential. DeepL Translator is applied to translate Chinese to English, and Tencent Translator is used to translate English to Chinese.

\section{Few-Shot on Unseen Downstream Datasets}
\begin{figure*}[h]
    \centering
    \includegraphics[page=1, width=1\textwidth]{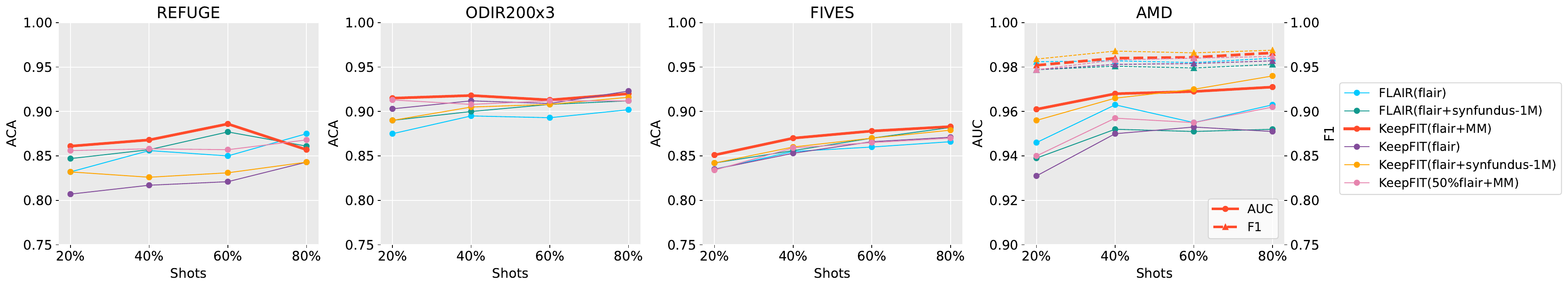}
    \caption{Few-shot experiments on unseen downstream datasets. }
    \label{fig:few-shot}
\end{figure*}

In this section, we present the supplementary results of Few-Shot experiments. The metric used for REFUGE, ODIR200$\times$3 and FIVES is ACA, and for AMD is AUC. From the Fig. ~\ref{fig:few-shot}, it is evident that KeepFit performs better with smaller training splits, like 20\%-60\%. As the training data increases, KeepFit's performance slightly declines, yet it consistently outperforms most baselines. This demonstrates its superior generalization capabilities and robustness.

\section{Disease and Abnormal Changes Categories in MM-Retinal}

This section provides the major categories of fundus diseases and abnormal changes in MM-Retinal with several example images in CFP, FFA and OCT modalities. These categories are summarized from the contents of four diagram books we used. 
\begin{figure}[H]
  \centering
  \includegraphics[page=1, width=1\textwidth]{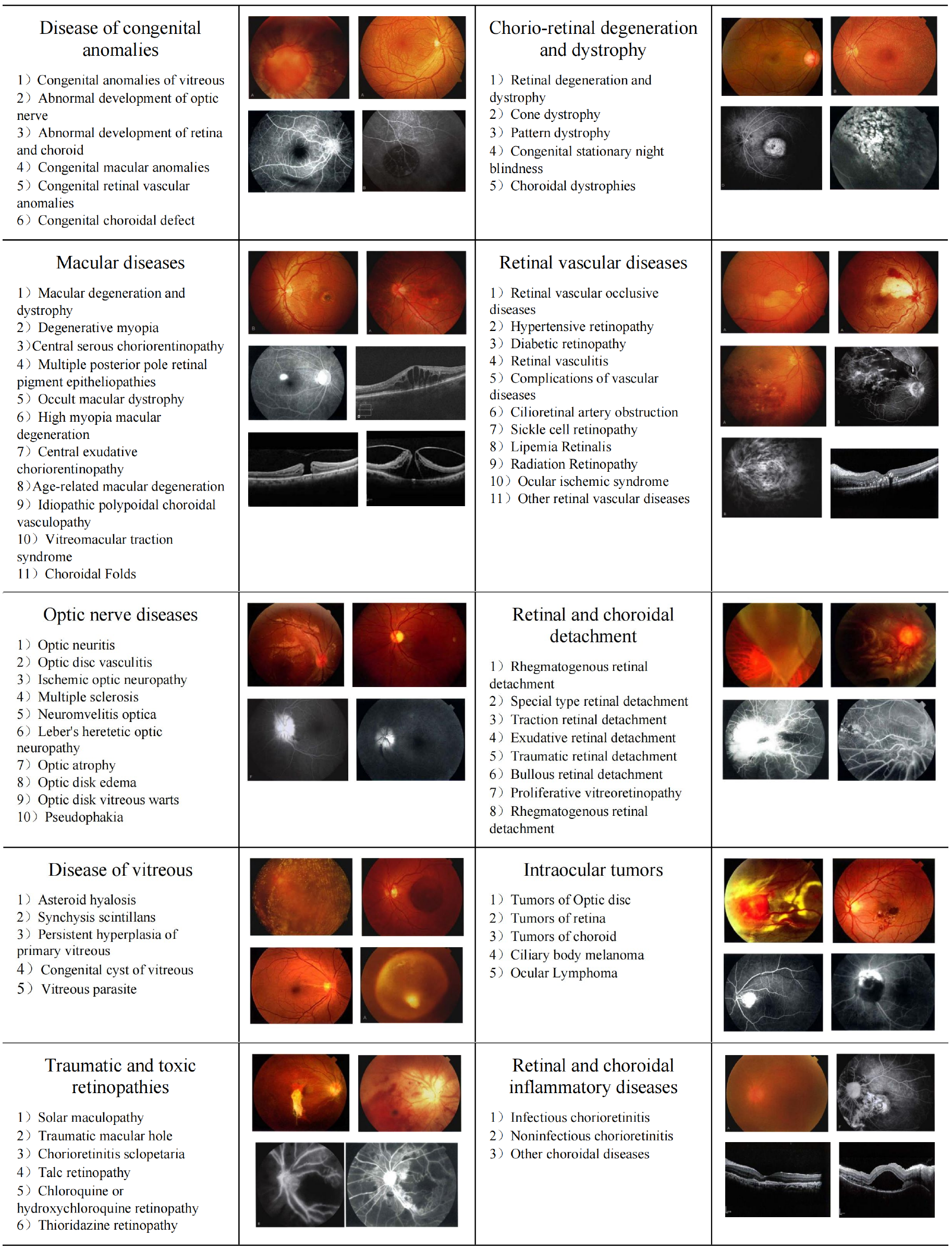}
  \caption{Major fundus diseases in MM-Retinal}
  \label{fig:mylabel}
\end{figure}

\begin{figure}[H]
  \includegraphics[page=1, width=1\textwidth]{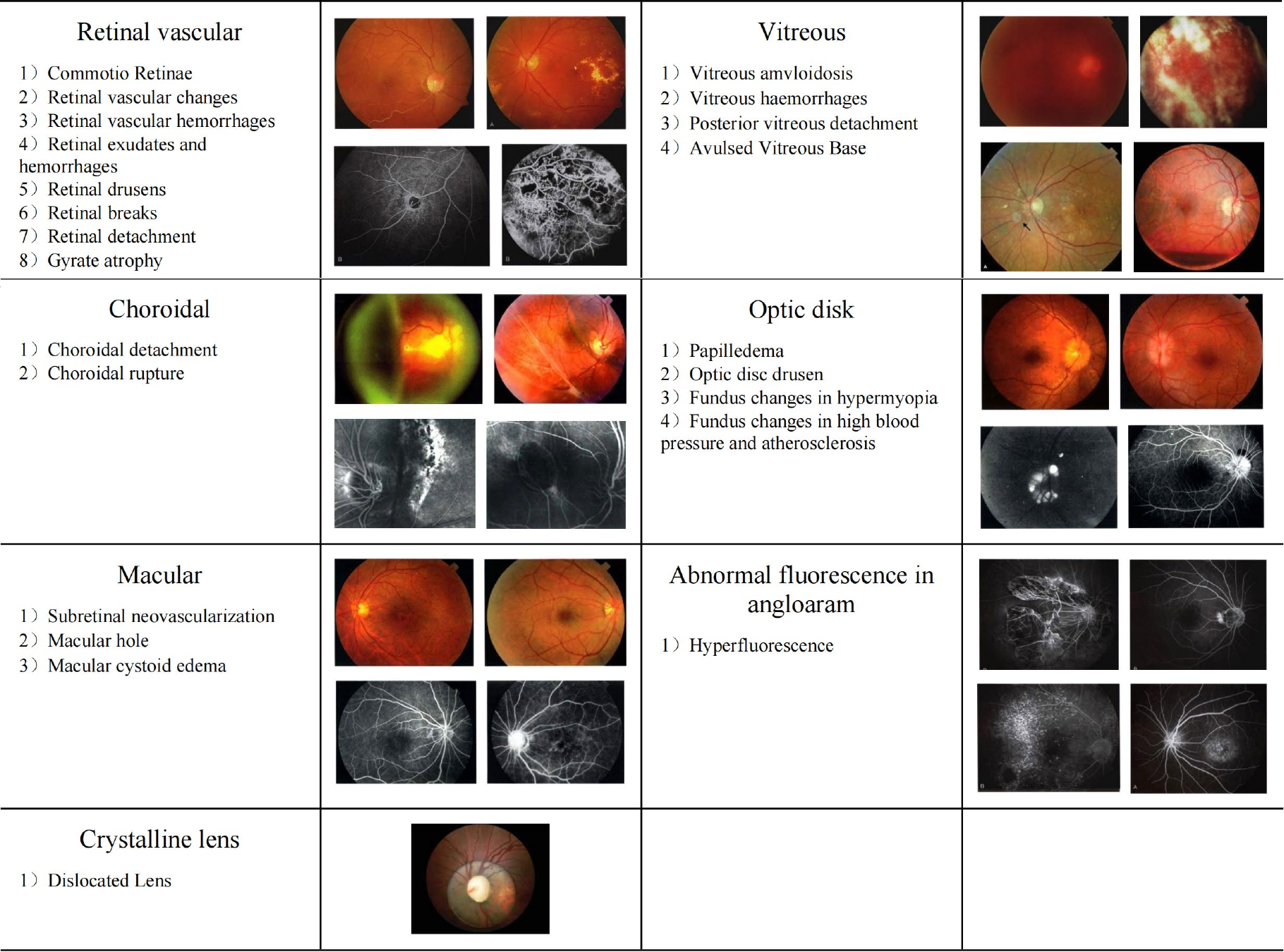}
  \caption{Major abnormal fundus changes in MM-Retinal}
  \label{fig:mylabel}
\end{figure}

\end{document}